\documentclass{article}

\usepackage{arxiv}

\usepackage[utf8]{inputenc} 
\usepackage[T1]{fontenc}    
\usepackage{url}            
\usepackage{booktabs}       
\usepackage{amsfonts}       
\usepackage{amsmath}
\usepackage{nicefrac}       
\usepackage{microtype}      
\usepackage{lipsum}
\usepackage{graphicx}
\graphicspath{ {./images/} }
\usepackage{multirow}

\usepackage{capt-of}   
\usepackage{etoolbox}
\AtBeginEnvironment{table}{\setlength{\belowcaptionskip}{5pt}}
\AtBeginEnvironment{table*}{\setlength{\belowcaptionskip}{5pt}}
\usepackage[table]{xcolor}
\definecolor{bgmanga}{HTML}{E3ECF7}
\definecolor{bgunseen}{HTML}{E2F2EE}
\definecolor{bgseen}{HTML}{ECE6F5}
\definecolor{bgfeat}{HTML}{F6E5E8}
\definecolor{bgabl}{HTML}{F7EEE2}
\definecolor{gainpos}{HTML}{2E7D32}
\definecolor{gainneg}{HTML}{C62828}
\colorlet{lmanga}{bgmanga!40}
\colorlet{lunseen}{bgunseen!40}
\colorlet{lseen}{bgseen!40}
\colorlet{lfeat}{bgfeat!40}
\colorlet{labl}{bgabl!40}

\newcommand{\grouprow}[3]{\rowcolor{#1}\multicolumn{#2}{l}{\textit{#3}}\\}
\newcommand{\ind}{\hspace{0.8em}}

\definecolor{cvprblue}{rgb}{0.21,0.49,0.74}
\usepackage[pagebackref,breaklinks,colorlinks,allcolors=cvprblue]{hyperref}
\usepackage[capitalize]{cleveref}
\crefname{section}{Sec.}{Secs.}
\Crefname{section}{Section}{Sections}
\crefname{table}{Tab.}{Tabs.}
\Crefname{table}{Table}{Tables}

\makeatletter
\renewenvironment{abstract}{%
  \centerline{\large\bfseries\scshape Abstract}\vspace{0.3em}\par}{\par\vspace{0.5em}}
\makeatother

\title{Where They Appear Tells Who They Are:\\
Book-Level Character Clustering in Manga}

\author{
  Siyuan Feng$^{1}$\quad
  Katsuhiko Hayashi$^{1}$\quad
  Hidetaka Kamigaito$^{2}$ \\[0.6em]
  \normalfont
  $^{1}$The University of Tokyo\quad
  $^{2}$Nara Institute of Science and Technology\\[0.4em]
  \small\texttt{9445233883@g.ecc.u-tokyo.ac.jp},\quad
  \texttt{katsuhiko-hayashi@g.ecc.u-tokyo.ac.jp},\quad
  \texttt{kamigaito.h@is.naist.jp}
}

\begin{document}

\twocolumn[{%
\renewcommand\twocolumn[1][]{#1}%
\maketitle
\begin{center}
  \centering
  \includegraphics[width=\textwidth]{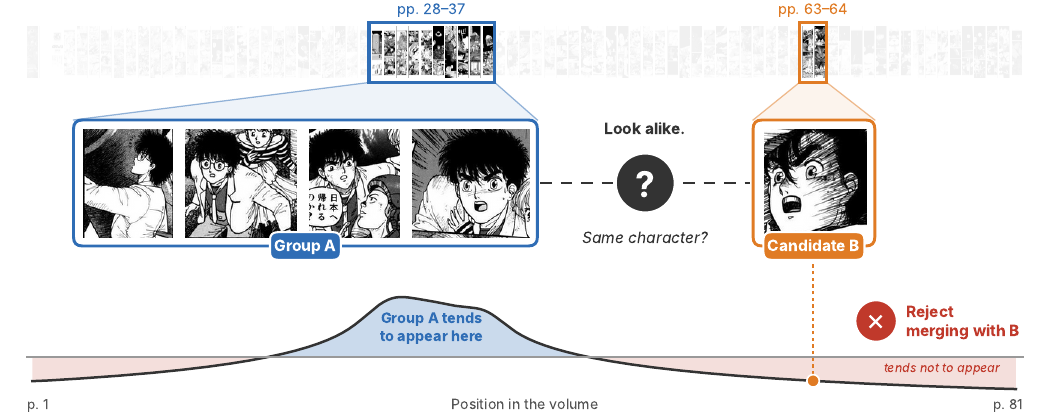}
  \captionof{figure}{
    Visual ambiguity and positional evidence in book-level
        character clustering.
        Representative instances from group A and candidate B
        have similar appearances but belong to different characters.
        Their positions within the volume provide additional evidence:
        B appears in a region where A has low relative presence,
        supporting rejection of the merge.
        The curve schematically illustrates relative presence;
        only representative instances are shown.
        Manga images are from ARMS \textcopyright Masaki Kato.
  }
  \label{fig:teaser}
\end{center}
\vspace{0.5em}
}]

\begin{abstract}
Grouping all appearances of the same character throughout a manga volume is a basic step toward story understanding.
We study this task, book-level manga character clustering, in which the number of characters is not known in advance.
Visual similarity alone is an unreliable cue: the same character may look substantially different across pages, while different characters may look alike.
We observe that a character who has appeared recently is likely to appear again nearby, so where characters appear across pages and panels provides identity evidence beyond appearance.
We propose presence-aware agglomeration, a trainable clustering method that progressively merges instances into groups, scoring each merge by visual similarity and group-level relative presence, which measures whether one group appears where the other is relatively active and is updated as groups grow.
On Manga109, our method achieves the best clustering performance at both the book and page levels.
It also improves clustering when transferred to another manga dataset without retraining and when combined with general-purpose visual encoders.
Code will be made publicly available.
\end{abstract}


\section{Introduction}
\label{sec:intro}

Associating character appearances throughout a manga volume supports
character-centric content organization and automated story
understanding~\cite{augereau2018survey,vivoli2024one}.
Book-level character clustering addresses this task by grouping character
instances across pages by identity, without requiring the number of
characters in advance.
Visual similarity is an ambiguous cue for this task: the same character
may look substantially different across pages, whereas different
characters may appear in similar poses, viewpoints, or expressions.
Even human readers inexperienced with visual narratives may interpret
successive depictions of the same person as distinct, similar-looking
individuals~\cite{cohn2020visual}.

Existing manga methods address identity association either within
individual pages~\cite{sachdeva2024manga,vivoli2024comix,soykan2023identity}
or across an entire
volume~\cite{tsubota2018adaptation,zhang2022unsupervised,sachdeva2024tails,baranwal2026re}.
At book scale, the difficulty grows: the same character may wear
different outfits or show different expressions in appearances dozens
of pages apart, while the number of other characters who happen to
match in pose or drawing style increases with the size of the volume.
A method must therefore connect different depictions of the same
character while distinguishing similar-looking characters.
Page and panel positions, which indicate where appearances occur in the volume, offer additional evidence for resolving such ambiguities.
However, most existing approaches rely primarily on visual
representations~\cite{zheng2020cartoon,zhang2023occlusion,soykan2023identity,nguyen2021manga},
and those that use positional information rely on manually specified
rules to construct training pairs or refine
pseudo-labels~\cite{tsubota2018adaptation,zhang2022unsupervised}.
Such rules apply the same notion of proximity or co-occurrence to every
character, even though characters differ greatly in how they appear
across a story.

Our starting intuition is that a character who has appeared recently is
more likely to appear again nearby.
Because some characters recur throughout a volume while others are
concentrated in particular portions of the story, the same position
provides different identity evidence for different characters.
As illustrated in Fig.~\ref{fig:teaser}, a candidate instance appearing
where a character group is relatively active supports their association,
whereas one appearing where the group is rarely present argues against it.
In this example, the candidate closely resembles the group in appearance
but appears on page 63--64, far from the pages where the group is active,
and indeed depicts a different character.
A single instance, however, reveals little about where its character
tends to appear; this information emerges only as instances of the same
character are grouped together.
We therefore estimate presence at the group level and update it as
groups grow, so that each association can draw on the appearance
locations accumulated by earlier ones.
We further measure presence relative to each group's own average
activity, so that frequently appearing main characters are not favored
simply because they appear everywhere.
We incorporate this evidence into \emph{presence-aware agglomeration},
a trainable clustering method that combines visual similarity with
dynamically updated presence and learns its merge-scoring function from
identity annotations, building on supervised hierarchical
clustering~\cite{yadav2019supervised}.

Experiments on Manga109~\cite{matsui2017sketch,multimedia_aizawa_2020}
show that our method outperforms existing methods and average-linkage clustering of frozen visual features at both the book and page levels, mainly by
separating visually similar characters that appearance-based clustering
merges.
Ablation studies confirm that each component contributes to this
improvement.
The method further transfers to
PopManga~\cite{sachdeva2024manga,sachdeva2024tails} without retraining
and improves clustering with general-purpose visual encoders such as
CLIP~\cite{radford2021learning} and DINOv2~\cite{oquab2023dinov2}.

Our contributions are twofold.
\begin{itemize}
    \item We show that where characters appear across the pages and
    panels of a manga volume provides identity evidence beyond visual
    appearance, and formulate it as group-level relative presence for
    book-level manga character clustering.

    \item We propose presence-aware agglomeration, a trainable
    clustering method that incorporates this evidence as character
    groups form, achieving the best performance among the evaluated
    methods on Manga109 and generalizing to other datasets and visual
    encoders.
\end{itemize}

\section{Related Work}
\label{sec:related_work}

\subsection{Manga Character Clustering}
\label{sec:rw_manga}

Computational analysis of comics has been reviewed in several
surveys~\cite{augereau2018survey,vivoli2024one}, and datasets such as
eBDtheque, COMICS, and
Manga109~\cite{guerin2013ebdtheque,iyyer2017amazing,matsui2017sketch,multimedia_aizawa_2020}
have supported character
detection~\cite{ogawa2018object,chu2017manga,sachdeva2024manga} and
character identification.
Most work on character identity focuses on visual representations,
including manga and cartoon face
representations~\cite{tsubota2018adaptation,zheng2020cartoon},
face--body combinations~\cite{zhang2022unsupervised,soykan2023identity},
occlusion-robust body features~\cite{zhang2023occlusion},
and multimodal features using visual and verbal
information~\cite{nguyen2021manga}; related ideas have also been
explored for animated characters~\cite{nir2022cast}.
These representations are associated at different scopes.
Some methods group characters within individual pages or short panel
sequences~\cite{sachdeva2024manga,vivoli2024comix,soykan2023identity},
whereas others operate across a book through book-specific adaptation,
reference character banks, or sequential identity
maintenance~\cite{tsubota2018adaptation,sachdeva2024tails,baranwal2026re}.
Other methods cluster characters in target books without identity labels,
either by applying density-based clustering to learned
features~\cite{yanagisawa2019manga,yanagisawa2020automatic} or through
pseudo-label learning~\cite{zhang2022unsupervised}.
Our work focuses on supervised book-level clustering, learning to
combine visual evidence with the distribution of character appearances
while determining identity groups without a reference character bank
or a predefined number of characters.

\subsection{Identity Clustering and Contextual Information in Other Domains}
\label{sec:rw_other_domains}

Identity clustering has been widely studied in
videos~\cite{everingham2006hello,wu2013constrained,zhang2016joint,jin2017end,otto2017clustering,tapaswi2019video,sharma2019self,kalogeiton2020constrained,brown2021face},
where changes in pose, illumination, occlusion, and scene context
complicate the grouping of recurring people.
These methods commonly exploit temporal structure, deriving must-link
and cannot-link constraints from face tracks and within-frame
co-occurrence~\cite{wu2013constrained,zhang2016joint,kalogeiton2020constrained,brown2021face},
learning track-level representations~\cite{sharma2019self}, or
estimating the number of identities during
clustering~\cite{jin2017end,tapaswi2019video}.
Person re-identification and person search similarly combine appearance
with spatial-temporal or visual-temporal
links~\cite{wang2019spatial,huang2018person}.
These cues do not transfer directly to manga.
Manga consists of discrete panels rather than continuous frames, so
there are no tracks linking consecutive appearances of a character,
and successive panels may jump across viewpoints, scenes, or story time.
Nor are there camera identities or timestamps from which movement
patterns could be estimated.
What manga does provide is the page and panel position of each
appearance.
Rather than converting these positions into fixed pairwise constraints,
we use them to estimate where each character group is relatively active
within a book and evaluate candidate identity associations against that
distribution.

\subsection{Clustering Methods and Learnable Agglomeration}
\label{sec:rw_agglomeration}

Clustering methods convert identity evidence into groups through
different mechanisms, including local feature
density~\cite{ester1996density,yanagisawa2019manga,yanagisawa2020automatic},
first-neighbor relations~\cite{sarfraz2019efficient,kalogeiton2020constrained},
and supervised graph-based linkage
prediction~\cite{yang2019learning,wang2019linkage,yang2020learning,shen2023star,xing2021learning}.
Hierarchical agglomerative clustering builds groups through successive
merges~\cite{mullner2011modern,otto2017clustering}, allowing group
membership to inform subsequent decisions, and supervised variants learn
the linkage used in this process~\cite{yadav2019supervised}.
This gradual construction is particularly suitable for modeling
character presence.
Each merge accumulates additional appearance locations, producing an
updated distribution that can inform the next association decision.
Our method follows this principle by learning a merge-scoring function
that combines visual similarity with group-level presence evidence.
As groups grow, their presence patterns and candidate merge scores are
updated together, allowing the clustering process to make increasing use
of the accumulated appearance positions.

\begin{figure*}[t]
    \centering
    \includegraphics[width=\textwidth]{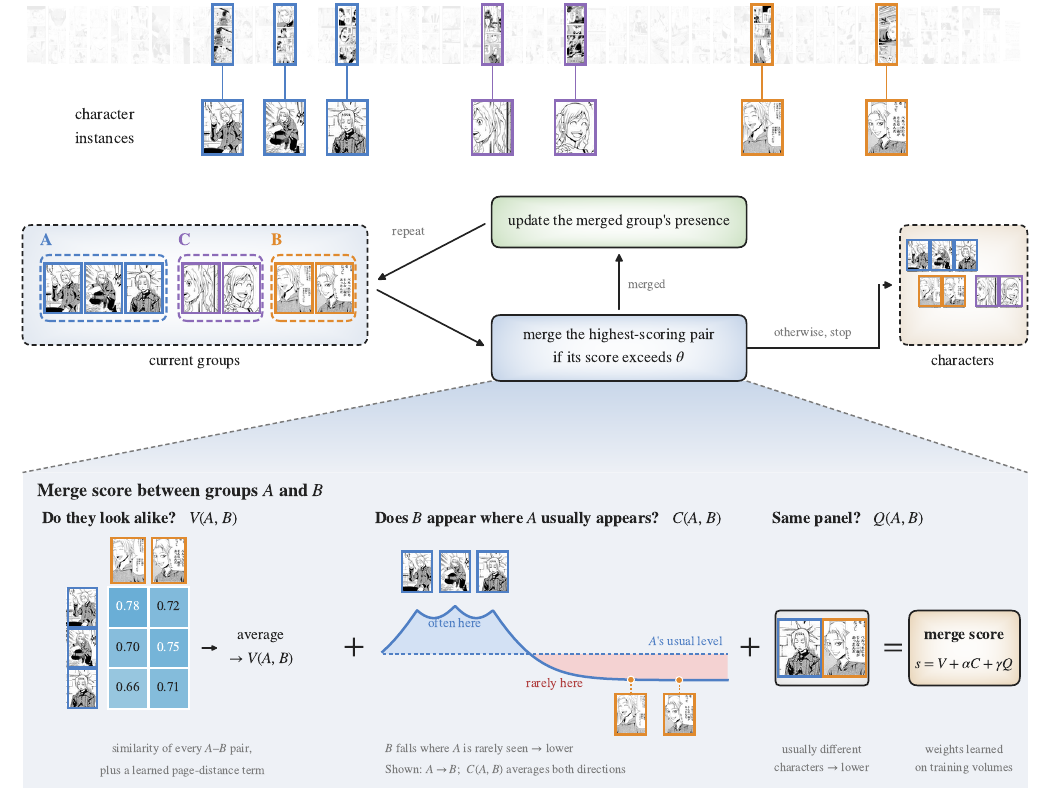}
    \caption{
        Overview of presence-aware agglomeration.
Starting from singleton groups, the method repeatedly merges
the highest-scoring pair whose score exceeds the threshold
$\theta$ and updates the merged group's presence pattern.
The merge score combines three terms:
$V(A,B)$, the cross-group average of visual similarity plus a learned
page-distance term $g$;
$C(A,B)$, the relative presence, which measures whether each group's
members appear where the other group is relatively active;
and $Q(A,B)$, the same-panel co-occurrence, which counts cross-group
instances appearing in the same panel.
The presence illustration shows the $A \rightarrow B$
direction; $C(A,B)$ averages both directions.
The displayed groups represent an intermediate clustering state; page positions, numerical values, and presence curves are illustrative. Character images are from Belmondo \textcopyright Shoei Ishioka.
    }
    \label{fig:method_overview}
\end{figure*}

\section{Method}
\label{sec:method}

\subsection{Problem Formulation and Overview}
\label{sec:method_overview}

Given a manga volume containing $N$ character body instances, we represent instance $i$ by $(x_i,p_i,f_i)$, where $x_i \in \mathbb{R}^{d}$ is its normalized visual feature, $p_i$ is its page index, and $f_i$ is its panel index in the book-wide reading sequence. Our goal is to partition these instances into identity groups without specifying the number of characters in advance. We use frozen visual features and learn how to combine visual similarity with page and panel positions. As illustrated in Fig.~\ref{fig:method_overview}, our method starts with individual instances and progressively merges them into groups. Each group's members provide evidence about where that character appears in the book. We summarize this evidence through a relative presence distribution and update it after each merge. Subsequent merge decisions therefore depend on both appearance similarity and the presence patterns accumulated by the current groups.

\subsection{Group-Level Relative Presence}
\label{sec:relative_presence}

We first define how strongly an appearance at one position supports an appearance at another, using transformed page and panel distances,
\begin{equation}
\begin{aligned}
u_{ij}^{p} &= \log(1+|p_i-p_j|), \\
u_{ij}^{f} &= \log(1+|f_i-f_j|),
\end{aligned}
\label{eq:position_distances}
\end{equation}
and a learnable positional kernel,
\begin{equation}
\kappa_{ij}
=
\exp\left[
h_{\phi}(u_{ij}^{p},u_{ij}^{f})
-a_pu_{ij}^{p}
-a_fu_{ij}^{f}
\right].
\label{eq:position_kernel}
\end{equation}
Here, $a_p$ and $a_f$ are learnable decay rates, and $h_{\phi}$ learns a correction to this relationship. Same-panel appearances require special treatment. Although their page and panel distances are both zero, distinct instances within a panel usually depict different characters. We therefore exclude same-panel pairs from proximity-based support,
\begin{equation}
K_{ij}
=
\mathbf{1}[f_i \neq f_j]\,\kappa_{ij},
\label{eq:masked_kernel}
\end{equation}
where $\mathbf{1}[\cdot]$ is the indicator function. Their contribution is modeled separately in the merge score. For a group $A$, we accumulate support from all its members at each original instance position $k$,
\begin{equation}
\begin{aligned}
\Lambda_A(k) &= \sum_{j \in A}K_{kj}, \\
L_A(k) &= \log\bigl(\epsilon+\Lambda_A(k)\bigr),
\end{aligned}
\label{eq:group_support}
\end{equation}
where $\epsilon>0$ is learnable. Characters have different appearance patterns across a volume, so we assess each position relative to the group's own average activity,
\begin{equation}
R_A(k)
=
L_A(k)
-\frac{1}{N}\sum_{r=1}^{N}L_A(r).
\label{eq:relative_presence}
\end{equation}
The average is taken over all original body-instance positions in the book. Thus, $R_A(k)>0$ indicates that position $k$ receives above-average support for group $A$. We evaluate two groups through their symmetric relative presence compatibility,
\begin{equation}
C(A,B)
=
\frac{1}{2}
\left[
\frac{1}{|B|}\sum_{j \in B}R_A(j)
+
\frac{1}{|A|}\sum_{i \in A}R_B(i)
\right].
\label{eq:presence_compatibility}
\end{equation}
This term asks whether each group's members appear where the other group is relatively active. It retains both positive and negative evidence, allowing the same position to support different identity associations depending on the groups' presence patterns.

\subsection{Presence-Aware Merge Scoring}
\label{sec:merge_scoring}

We combine relative presence with visual similarity and a general page-distance relationship. The score between instances is
\begin{equation}
v_{ij}
=
x_i^{\top}x_j+g_{\psi}(u_{ij}^{p}),
\label{eq:instance_score}
\end{equation}
where $g_{\psi}$ is a learnable function shared across characters and books. We aggregate these scores through their cross-group average,
\begin{equation}
V(A,B)
=
\frac{1}{|A||B|}
\sum_{i \in A}\sum_{j \in B}v_{ij}.
\label{eq:group_pair_score}
\end{equation}
To account explicitly for the same-panel exception to proximity-based association, we define
\begin{equation}
Q(A,B)
=
\frac{
\displaystyle
\sum_{i \in A}\sum_{j \in B}\mathbf{1}[f_i=f_j]
}{
\min(|A|,|B|)
}.
\label{eq:cooccurrence}
\end{equation}
This term measures the number of cross-group same-panel pairs relative to the smaller group's size. The complete merge score is
\begin{equation}
s(A,B)
=
V(A,B)+\alpha C(A,B)+\gamma Q(A,B),
\label{eq:merge_score}
\end{equation}
where $\alpha$ and $\gamma$ are learnable coefficients with unrestricted signs. A negative $\gamma$ reduces the merge score when group members occur in the same panel, while allowing other evidence to support a merge when one character is depicted repeatedly within a panel. The general term $g_{\psi}$ accounts for how the reliability of visual similarity as identity evidence varies with page distance, since both the drawing of the same character and the chance resemblance between different characters change across a volume. The relative presence term $C(A,B)$ further considers where the current groups are active within the book. The two terms express the same intuition at different levels: $g_{\psi}$ captures how identity evidence varies with page distance on average over all characters, whereas $C(A,B)$ applies this idea to each group's own appearance record. Together, these components combine instance-level appearance evidence with group-dependent positional information.

\subsection{Learning the Merge Function}
\label{sec:training}

We train the model using supervision at both the instance-pair and group-merge levels. The general term $g_{\psi}$ describes how the reliability of visual similarity varies with page distance, a relationship that holds for instance pairs at all distances. We therefore learn it from all eligible instance pairs in a book. Merge candidates, in contrast, are high-scoring group pairs selected during agglomeration and cover only a biased subset of page distances, but they reflect the decisions that the group-level terms must make. We therefore learn the presence and co-occurrence terms from merge candidates.

Both objectives use weighted binary cross-entropy, written as $\operatorname{BCE}_{w}(q,y)$, where $q$ is the predicted probability, $y \in \{0,1\}$ is the identity-association label, and $w$ weights positive examples to account for class imbalance. For instance-pair learning, we construct a set $\mathcal{D}_{\mathrm{pair}}$ of labeled body-instance pairs. A pair receives label $y_{ij}=1$ when both instances belong to the same individually identified character, and $y_{ij}=0$ when they belong to different characters. The dataset also contains an \emph{Other} category for instances without individual character names, typically passersby or background characters. This category contains visually diverse individuals whose shared label does not establish a common identity. We therefore exclude Other--Other pairs and use identified-character--Other pairs as negatives. We convert the instance-pair score $v_{ij}$ into a probability using the sigmoid function $\sigma$, a learnable positive scale $\beta_{\mathrm{pair}}$, and a learnable offset $b_{\mathrm{pair}}$. The resulting objective is
\begin{equation}
\begin{aligned}
\mathcal{L}_{\mathrm{pair}}
&=
\frac{1}{|\mathcal{D}_{\mathrm{pair}}|}
\sum_{(i,j)\in\mathcal{D}_{\mathrm{pair}}}
\operatorname{BCE}_{w_{\mathrm{pair}}}
\left(
q_{ij}^{\mathrm{pair}},y_{ij}
\right), \\
q_{ij}^{\mathrm{pair}}
&=
\sigma\left(
\beta_{\mathrm{pair}}v_{ij}+b_{\mathrm{pair}}
\right).
\end{aligned}
\label{eq:pair_loss}
\end{equation}

For group-level learning, we generate a candidate set $\mathcal{D}_{\mathrm{merge}}$ through teacher-constrained agglomeration. Starting from singleton groups, the teacher repeatedly performs the highest-scoring merge under the current model between groups belonging to the same identified character. At intermediate stages, we collect high-scoring candidate group pairs. A candidate receives label $y_{AB}=1$ when its groups belong to the same character and $y_{AB}=0$ otherwise. Each Other instance is treated as a separate singleton for this supervision, making candidate merges involving it negative. With the sampled group memberships fixed, we recompute the merge scores $s(A,B)$. A learnable positive scale $\beta_{\mathrm{merge}}$ and a learnable offset $\tau_{\mathrm{train}}$ convert these scores into association probabilities,
\begin{equation}
\begin{aligned}
\mathcal{L}_{\mathrm{merge}}
&=
\frac{1}{|\mathcal{D}_{\mathrm{merge}}|}
\sum_{(A,B)\in\mathcal{D}_{\mathrm{merge}}}
\operatorname{BCE}_{w_{\mathrm{merge}}}
\left(
q_{AB}^{\mathrm{merge}},y_{AB}
\right), \\
q_{AB}^{\mathrm{merge}}
&=
\sigma\left(
\beta_{\mathrm{merge}}
\left[s(A,B)-\tau_{\mathrm{train}}\right]
\right).
\end{aligned}
\label{eq:merge_loss}
\end{equation}
The weights $w_{\mathrm{pair}}$ and $w_{\mathrm{merge}}$ are the negative-to-positive example ratios in their respective sets within each training book. We combine the objectives as
\begin{equation}
\mathcal{L}
=
\mathcal{L}_{\mathrm{pair}}
+
\mathcal{L}_{\mathrm{merge}}.
\label{eq:total_loss}
\end{equation}

The pair objective updates the page-distance function $g_{\psi}$ and its probability calibration parameters. When computing the merge objective, we hold $V(A,B)$ constant with respect to differentiation, so its gradients update the positional kernel, $\epsilon$, the coefficients $\alpha,\gamma$, and the merge calibration parameters. This prevents the biased distribution of merge candidates from altering the general page-distance term, which remains supervised only by instance pairs. This division learns general distance evidence from instance pairs and group-dependent presence evidence from candidate merges. All instances, including Other instances, remain available when computing presence distributions and participate in inference. Evaluation measures identity clustering on instances with individual character identities.

\subsection{Agglomerative Inference}
\label{sec:inference}

We initialize each instance as a singleton group,
\begin{equation}
\mathcal{P}_0
=
\bigl\{\{1\},\ldots,\{N\}\bigr\}.
\label{eq:initial_partition}
\end{equation}
At each step, we select the highest-scoring pair among all current groups,
\begin{equation}
(A_t,B_t)
=
\underset{
\substack{
A,B\in\mathcal{P}_{t-1}\\
A\neq B
}
}{\arg\max}
s(A,B),
\label{eq:greedy_merge}
\end{equation}
and merge them into $U_t=A_t\cup B_t$. The merged group accumulates support from both constituent groups,
\begin{equation}
\Lambda_{U_t}(k)
=
\Lambda_{A_t}(k)+\Lambda_{B_t}(k).
\label{eq:support_update}
\end{equation}
We then update its relative presence distribution and its scores against the remaining groups. These updates are incremental: since $V$ is an average of instance-pair scores, $V(U_t,X)$ is the size-weighted average of $V(A_t,X)$ and $V(B_t,X)$ for any remaining group $X$, and the numerator of $Q$ is likewise additive, so only $C$ requires recomputing the relative presence of the merged group. In this way, the growing collection of appearance locations informs later merge decisions. We record the full merge sequence with scores $m_1,\ldots,m_{N-1}$. Because group-level evidence changes after merging, later scores may increase. We therefore use the historical minimum,
\begin{equation}
\bar{m}_t
=
\min_{1\leq r\leq t}m_r,
\label{eq:historical_minimum}
\end{equation}
to obtain a consistent prefix of the sequence. Given an inference threshold $\theta$, the final partition follows the first $T^{*}$ merges,
\begin{equation}
T^{*}
=
\max\left(
\{0\}\cup\{t:\bar{m}_t>\theta\}
\right).
\label{eq:partition_cut}
\end{equation}
The threshold and model checkpoint are selected using mean book-level AMI on validation books. This procedure determines the number of identity groups and produces a consistent partition across the volume. The same partition can also be restricted to individual pages for page-level evaluation.

\begin{table*}[t]
\centering
\small
\setlength{\tabcolsep}{5pt}
\setlength{\aboverulesep}{0pt}
\setlength{\belowrulesep}{0pt}
\renewcommand{\arraystretch}{1.18}
\caption{
Character clustering on Manga109, transfer to PopManga, visual-feature
substitution, and component ablations.
Unnamed ground-truth instances are excluded from scoring.
Page-level scores restrict each predicted volume or chapter partition
to individual pages and are averaged over all valid pages, including
single-instance pages.
Magi clusters each page independently, so its book/chapter-level scores
are not reported.
Average-linkage clustering uses cosine similarities of frozen MagiV2
features; all models and thresholds used on PopManga are selected on
Manga109.
In feature substitution, each feature row applies average-linkage
clustering to frozen CLIP or DINOv2 features, and + Ours applies our
agglomeration to the same features.
Ablations retrain the model after one change: w/o $g$ removes the
page-distance term inside $V$, w/o $C$ removes relative presence,
w/o $Q$ removes the same-panel term, and Absolute $C$ uses absolute
rather than relative presence.
NMI uses arithmetic normalization.
Subscripts report sample standard deviations over three seeds.
Bold marks the best score in each main-result block.
}
\label{tab:main_results}
\begin{tabular}{lcccccc}
\toprule
\multirow{2}{*}{Method}
& \multicolumn{4}{c}{Book/chapter-level}
& \multicolumn{2}{c}{Page-level} \\
\cmidrule(lr){2-5}
\cmidrule(lr){6-7}
& AMI $\uparrow$ & NMI $\uparrow$ & ARI $\uparrow$
& B$^3$ F1 $\uparrow$ & AMI $\uparrow$ & NMI $\uparrow$ \\
\midrule
\grouprow{bgmanga}{7}{Manga109}
\rowcolor{lmanga}\ind Magi
& -- & -- & -- & -- & $0.602$ & $0.748$ \\
\rowcolor{lmanga}\ind MagiV2
& $0.515$ & $0.548$ & $0.432$ & $0.507$ & $0.626$ & $0.756$ \\
\rowcolor{lmanga}\ind Average-linkage clustering
& $0.631$ & $0.663$ & $0.544$ & $0.629$ & $0.684$ & $0.797$ \\
\rowcolor{lmanga}\ind \textbf{Ours}
& $\mathbf{0.699}_{\pm0.004}$ & $\mathbf{0.727}_{\pm0.003}$ & $\mathbf{0.601}_{\pm0.006}$ & $\mathbf{0.677}_{\pm0.001}$ & $\mathbf{0.754}_{\pm0.002}$ & $\mathbf{0.836}_{\pm0.001}$ \\
\midrule
\grouprow{bgunseen}{7}{PopManga-unseen}
\rowcolor{lunseen}\ind Magi
& -- & -- & -- & -- & $0.703$ & $0.814$ \\
\rowcolor{lunseen}\ind MagiV2
& $0.584$ & $0.648$ & $0.609$ & $0.704$ & $0.683$ & $0.811$ \\
\rowcolor{lunseen}\ind Average-linkage clustering
& $0.682$ & $0.709$ & $0.685$ & $0.774$ & $0.708$ & $0.808$ \\
\rowcolor{lunseen}\ind \textbf{Ours}
& $\mathbf{0.707}_{\pm0.003}$ & $\mathbf{0.733}_{\pm0.002}$ & $\mathbf{0.693}_{\pm0.005}$ & $\mathbf{0.775}_{\pm0.004}$ & $\mathbf{0.751}_{\pm0.002}$ & $\mathbf{0.841}_{\pm0.002}$ \\
\midrule
\grouprow{bgseen}{7}{PopManga-seen}
\rowcolor{lseen}\ind Magi
& -- & -- & -- & -- & $0.691$ & $0.800$ \\
\rowcolor{lseen}\ind MagiV2
& $0.541$ & $0.613$ & $0.572$ & $0.672$ & $0.667$ & $0.798$ \\
\rowcolor{lseen}\ind Average-linkage clustering
& $0.661$ & $0.691$ & $0.675$ & $0.756$ & $0.707$ & $0.810$ \\
\rowcolor{lseen}\ind \textbf{Ours}
& $\mathbf{0.687}_{\pm0.004}$ & $\mathbf{0.715}_{\pm0.004}$ & $\mathbf{0.688}_{\pm0.002}$ & $\mathbf{0.770}_{\pm0.000}$ & $\mathbf{0.744}_{\pm0.006}$ & $\mathbf{0.833}_{\pm0.005}$ \\
\midrule
\grouprow{bgfeat}{7}{Feature substitution (Manga109)}
\rowcolor{lfeat}\ind CLIP features
& $0.263$ & $0.378$ & $0.135$ & $0.303$ & $0.187$ & $0.524$ \\
\rowcolor{lfeat}\ind \ind + Ours
& $0.378_{\pm0.006}$ & $0.466_{\pm0.005}$ & $0.154_{\pm0.005}$ & $0.263_{\pm0.006}$ & $0.367_{\pm0.013}$ & $0.555_{\pm0.013}$ \\
\rowcolor{lfeat}\ind DINOv2 features
& $0.142$ & $0.307$ & $0.061$ & $0.316$ & $0.087$ & $0.431$ \\
\rowcolor{lfeat}\ind \ind + Ours
& $0.313_{\pm0.003}$ & $0.366_{\pm0.013}$ & $0.148_{\pm0.009}$ & $0.322_{\pm0.023}$ & $0.094_{\pm0.045}$ & $0.222_{\pm0.086}$ \\
\midrule
\grouprow{bgabl}{7}{Ablations (Manga109)}
\rowcolor{labl}\ind w/o $g$
& $0.654_{\pm0.001}$ & $0.681_{\pm0.002}$ & $0.545_{\pm0.007}$ & $0.643_{\pm0.003}$ & $0.722_{\pm0.003}$ & $0.814_{\pm0.006}$ \\
\rowcolor{labl}\ind w/o $C$
& $0.670_{\pm0.019}$ & $0.698_{\pm0.021}$ & $0.569_{\pm0.037}$ & $0.660_{\pm0.016}$ & $0.724_{\pm0.002}$ & $0.817_{\pm0.005}$ \\
\rowcolor{labl}\ind w/o $Q$
& $0.674_{\pm0.007}$ & $0.698_{\pm0.008}$ & $0.553_{\pm0.012}$ & $0.655_{\pm0.008}$ & $0.705_{\pm0.013}$ & $0.794_{\pm0.014}$ \\
\rowcolor{labl}\ind Absolute $C$
& $0.693_{\pm0.003}$ & $0.721_{\pm0.003}$ & $0.576_{\pm0.004}$ & $0.676_{\pm0.005}$ & $0.743_{\pm0.007}$ & $0.825_{\pm0.007}$ \\
\bottomrule
\end{tabular}
\end{table*}

\section{Experiments}
\label{sec:exp}

\subsection{Experimental Setup}
\label{sec:experimental_setup}

We evaluate book-level character clustering on Manga109~\cite{matsui2017sketch,multimedia_aizawa_2020} and assess cross-dataset transfer on PopManga~\cite{sachdeva2024manga,sachdeva2024tails}. Following the Manga109 split used in prior work~\cite{tsubota2018adaptation}, we use 83 volumes for training, 10 for validation, and 11 for testing. Our inputs consist of annotated character body boxes, frozen MagiV2 body features~\cite{sachdeva2024tails}, page indices, and panel positions in the reading sequence estimated with a panel order estimator~\cite{kovanen2015layered}. All body instances participate in clustering, including those annotated as Other. Evaluation excludes Other instances and measures the identity assignments of individually identified characters. For transfer, we evaluate the 20 unseen and 30 seen chapters of PopManga separately; the seen and unseen splits indicate whether the series were included in Magi's training data. Our clustering model is trained only on Manga109, and its checkpoint and inference threshold are selected using Manga109 validation books.

We report book-level AMI, NMI, ARI, and B$^3$ F1. Book-level scores are computed separately for each volume and averaged across volumes. For page-level evaluation, we restrict the predicted book-level partition to each page and compute AMI and NMI using that page's predicted and ground-truth identity labels. We then average the scores over all valid pages in the evaluation set, giving each page equal weight. PopManga follows the corresponding chapter-level procedure, with chapter-level scores averaged across chapters and page-level scores averaged across valid pages. Trainable variants are run with three random seeds, and we report their mean and sample standard deviation. Deterministic methods and fixed-reference character-bank configurations are evaluated once. Training uses AdamW with a learning rate of $3\times10^{-4}$, weight decay of $10^{-4}$, and cosine learning-rate decay. We train for up to 50 epochs, with early stopping after 10 epochs without improvement in validation book-level AMI. Detailed architecture, initialization, sampling, and threshold-selection settings are provided in the appendix.

We compare with the following methods. Magi provides the page-level character-clustering comparison; we evaluate its original checkpoint under our protocol, using a fixed affinity threshold of $0.65$ and connected components on each page. MagiV2 uses a reference character bank to obtain book-level identity assignments, which are also projected onto pages. On Manga109, we evaluate its stored predictions from a fixed reference configuration, using one ground-truth body crop per individually identified character as a reference and excluding reference crops from the query set. On PopManga, we use the supplied reference character banks. We further include average-linkage clustering of the same frozen MagiV2 body features using cosine similarity, with its threshold selected on Manga109 validation books, as a direct visual comparison with our method.

\subsection{Main Results}
\label{sec:main_results}

\begin{figure*}[t]
    \centering
    \includegraphics[width=\textwidth]{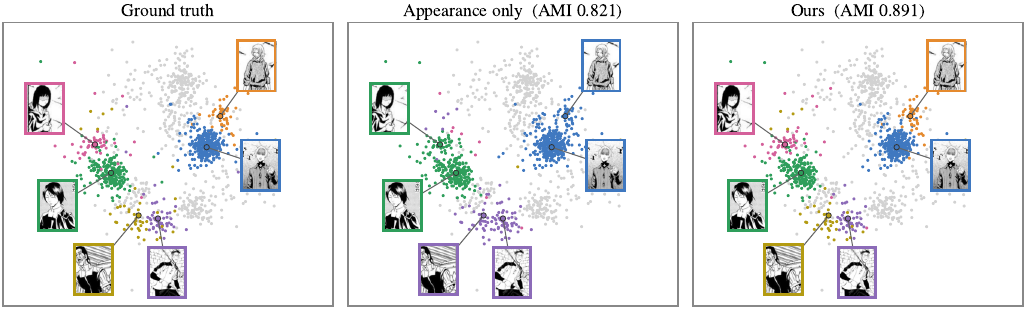}
    \caption{
        Qualitative comparison on a test manga volume.
        From left to right: ground-truth identities,
        average-linkage clustering, and ours.
        All panels share the same two-dimensional coordinates.
        Colors indicate identities or predicted clusters;
        selected identities are highlighted, with the remaining
        instances shown in gray.
        Representative character crops illustrate visually similar
        identities that average-linkage clustering merges,
        while our method separates them.
        Character images are from Belmondo \textcopyright Shoei Ishioka.
    }
    \label{fig:qualitative_comparison}
\end{figure*}

Table~\ref{tab:main_results} groups the results into main results on Manga109, transfer to PopManga, visual-feature substitution, and component ablations. On Manga109, our method achieves the best scores on all book-level and page-level metrics. The most informative comparison is with average-linkage clustering: it uses the same frozen features and the same cross-group averaging of visual similarity as our agglomeration, differing only in the absence of positional terms. The improvement of $0.068$ in book-level AMI can therefore be attributed to the page-distance and group-level presence evidence. MagiV2's reference-bank assignments score below both clustering methods in the evaluated configuration, even though they require a reference image for every character, which our method does not. When restricted to individual pages, our book-level partition also achieves higher page-level scores than Magi, which clusters each page independently.

Figure~\ref{fig:qualitative_comparison} illustrates where the improvement comes from. Since the features are fixed, all panels share the same coordinates and differ only in their cluster assignments. Average-linkage clustering merges three pairs of visually similar characters into single clusters, whereas our method separates all three pairs. In these cases, visual similarity alone is insufficient, and the positions at which the characters appear provide the evidence needed to keep them apart. Per-volume results and additional qualitative examples are provided in the appendix.

Earlier studies report results under different evaluation conditions. AMFR achieves an NMI of $0.71$ by clustering face instances with the ground-truth character count supplied and adapting its representation to each target volume~\cite{tsubota2018adaptation}. Manga-MMTL reports an AMI of $0.6791$ for its M3+M5 feature combination, using face images and panel-associated text while excluding characters with fewer than ten appearances per book~\cite{nguyen2021manga}. Our evaluation covers all individually identified body instances, including infrequent characters, and determines the number of identities during clustering using a model fixed at test time. These differences prevent a direct numerical comparison with our results.

\subsection{Ablation Studies}
\label{sec:ablation_studies}

The ablation block of Table~\ref{tab:main_results} retrains the model after removing or changing one component, with its checkpoint and threshold selected on the validation set. All four modifications reduce book-level AMI. Removing the general page-distance term $g$ causes the largest drop, to $0.654$, indicating that adjusting visual similarity by page distance is important even before group-level evidence is considered. Removing the relative presence term $C$ reduces book-level AMI to $0.670$ and markedly increases the variance across seeds, with a standard deviation of $0.019$ compared with $0.004$ for the full model, suggesting that presence evidence also stabilizes merge decisions.

Removing the same-panel term $Q$ causes the largest page-level drop among the variants, from $0.754$ to $0.705$. This is consistent with its role: instances in the same panel necessarily lie on the same page, so incorrectly merging characters depicted together directly harms within-page grouping. Finally, replacing the relative presence $R_A$ with the uncentered log support $L_A$ in Eq.~\eqref{eq:presence_compatibility} leads to a smaller but consistent reduction, most visible at the page level. Uncentered support also reflects a group's overall activity and thus favors frequently appearing characters, whereas centering asks whether a position is particularly supported for that group. The appendix further reports the learned page-distance term and coefficients, the sensitivity to the inference threshold $\theta$, and additional positional and training-objective analyses.

\subsection{Generalization Across Datasets and Visual Encoders}
\label{sec:generalization}

On PopManga, models and thresholds selected on Manga109 improve over average-linkage clustering on both splits, outperform MagiV2's reference banks on all chapter-level metrics, and yield the highest page-level scores. The gains are similar on the seen and unseen splits, indicating that the benefit of presence modeling does not depend on whether the visual encoder was trained on the series. They are smaller than on Manga109, which may reflect the shorter chapter units of PopManga: with fewer pages per chapter, there is less positional context from which group presence can be estimated.

Replacing the frozen body features with CLIP~\cite{radford2021learning} or DINOv2~\cite{oquab2023dinov2} features and retraining the clustering model on Manga109 also improves book-level AMI, from $0.263$ to $0.378$ and from $0.142$ to $0.313$, respectively. These gains are larger than with MagiV2 features, suggesting that positional evidence compensates more when visual evidence is weaker, although MagiV2 remains the strongest representation overall. The effects are less uniform on the complementary metrics. With CLIP, B$^3$ F1 decreases despite the AMI improvement, and with DINOv2, page-level scores vary considerably across seeds and page-level NMI decreases, indicating less stable within-page assignments. The feature-substitution experiments therefore demonstrate consistent book-level AMI gains across encoders, while revealing encoder-dependent differences in cluster structure.

\section{Conclusion}
\label{sec:conclusion}

We presented presence-aware agglomeration for book-level manga character
clustering. Our method uses page and panel positions to model where each
character group is relatively active within a volume, providing identity
evidence that complements visual similarity. As groups are merged, their
accumulated appearance locations update the presence patterns used in
subsequent decisions. Experiments on Manga109 show improvements at both
the book and page levels, and transfer to PopManga and experiments with
alternative visual encoders demonstrate applicability across datasets
and representations.
These results indicate that where characters appear is a useful and
transferable source of identity evidence in manga, complementary to
improvements in visual representations.
Our experiments use annotated character boxes; evaluating the method
with detected characters is a natural next step toward fully automatic
use.
In addition, the current method makes irreversible merge decisions,
allowing early errors to persist. Future work could explore revisable
clustering that uses evolving presence evidence to reconsider earlier
associations.

\bibliographystyle{unsrt}
\bibliography{references}

\appendix
\twocolumn[{%
  \centering
  {\Large\bfseries Appendix\par}
  \vspace{1.5em}
}]
\section{Implementation Details}
\label{sec:supp_impl}

\paragraph{Architecture.}
The page-distance function $g_{\psi}$ and the kernel correction $h_{\phi}$ are multilayer perceptrons with one hidden layer of 16 GELU units, taking $u^{p}_{ij}$ and $(u^{p}_{ij},u^{f}_{ij})$ as input, respectively.
The decay rates $a_p$ and $a_f$ and the support floor $\epsilon$ are parameterized with softplus to remain positive, whereas $\alpha$ and $\gamma$ are not sign-constrained.

\paragraph{Initialization.}
The output layers of $g_{\psi}$ and $h_{\phi}$ are initialized to zero, so both functions output zero at initialization; the other layers use the default PyTorch initialization.
The decay rates and $\epsilon$ are initialized to 1, so the initial kernel already decreases with distance.
The coefficient $\alpha$ is initialized to 0.3 and $\gamma$ to 0.
The calibration parameters $\beta_{\mathrm{pair}}$, $b_{\mathrm{pair}}$, $\beta_{\mathrm{merge}}$, and $\tau_{\mathrm{train}}$ are initialized from a logistic regression of instance-pair labels on visual similarity, fitted on the training books.

\paragraph{Sampling.}
The pair objective uses all eligible unordered instance pairs in each training volume, without subsampling.
For the merge objective, candidate pairs are collected at ten evenly spaced stages of the teacher-constrained agglomeration.
At each stage, every current group contributes its five highest-scoring partners, and duplicate pairs within a stage are removed.
Each parameter update uses one complete training volume, and the volume order is shuffled in every epoch.

\paragraph{Threshold selection.}
Candidate thresholds are 200 quantiles, from 0.005 to 0.995, of the merge scores pooled over the validation volumes, together with two values just outside the minimum and maximum scores.
For each validation epoch, the threshold with the highest mean book-level AMI is selected, and the checkpoint is chosen by the same criterion.

\section{Evaluation Details}
\label{sec:supp_eval}

\paragraph{Metrics.}
AMI measures the mutual information between the predicted and ground-truth partitions, corrected for the agreement expected by chance.
NMI normalizes the mutual information by the arithmetic mean of the entropies of the two partitions.
ARI measures the agreement on whether pairs of instances belong to the same cluster, also corrected for chance.
B$^3$ F1 is the harmonic mean of B$^3$ precision and recall, which are computed for each instance from the overlap between its predicted cluster and its ground-truth identity and then averaged over instances.

\begin{figure}[t]
\centering
\includegraphics[width=\linewidth]{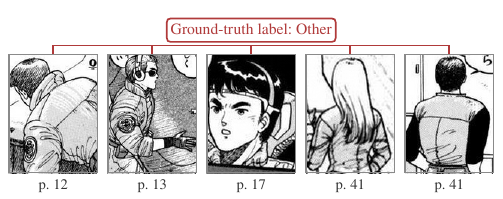}
\caption{Instances annotated as Other in one Manga109 volume. They depict different individuals but share the same label, and they are therefore excluded from evaluation. Manga images are from ARMS \textcopyright Masaki Kato.}
\label{fig:supp_other}
\end{figure}

\paragraph{Other instances and predictions.}
\Cref{fig:supp_other} shows examples of instances annotated as Other in a single volume.
They depict different individuals, such as passersby and unnamed crew members, but share the same label, so treating the label as an identity would reward merging unrelated people.
On PopManga, an instance is treated as Other when its character name is missing or empty, which corresponds to the Other label in Manga109.
MagiV2 can also assign instances to its own Other category.
For evaluation, these instances are placed in one shared cluster on Manga109 and in separate singleton clusters on PopManga.

\section{Learned Parameters and Threshold Sensitivity}
\label{sec:supp_params}

\begin{figure}[t]
\centering
\includegraphics[width=\linewidth]{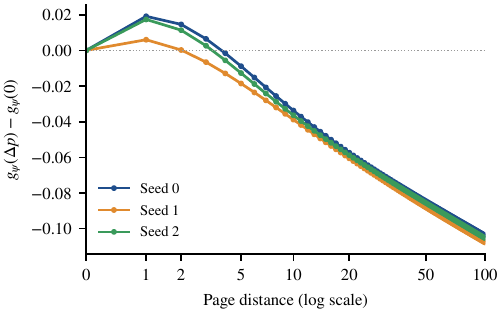}
\caption{Learned page-distance term $g_{\psi}$ of the three models, shown relative to its value at zero page distance. The horizontal axis uses a logarithmic scale.}
\label{fig:supp_page}
\end{figure}

\paragraph{Page-distance term.}
\Cref{fig:supp_page} shows the learned page-distance term $g_{\psi}$ relative to its value at zero page distance.
In all three models, $g_{\psi}$ first increases slightly from zero to one page (by 0.006--0.019), returns to its zero-distance value at about three to four pages, and then decreases almost linearly in the logarithm of page distance, reaching about $-0.10$ at 100 pages.
Visual similarity is thus discounted increasingly as instances lie farther apart in the volume.

\paragraph{Coefficients.}
\Cref{tab:supp_params} lists the selected epoch, learned coefficients, and validation-selected threshold of each model.
The values are consistent across seeds, with $\alpha\approx0.20$ and $\gamma$ between $-0.12$ and $-0.10$.
The negative $\gamma$ lowers the merge score of groups whose members share panels, in line with the role of the same-panel term.
The decay rates $a_p$ and $a_f$ remain close to 1, and the positional kernel decreases rapidly with distance.
For example, $\kappa$ is 0.089 for instances three panels apart on the same page, 0.0074 for instances eleven panels apart on adjacent pages, and $2.3\times10^{-4}$ for instances fifty panels apart across five pages, averaged over the three models.
Presence support is therefore determined mainly by nearby appearances.

\begin{figure}[t]
\centering
\includegraphics[width=\linewidth]{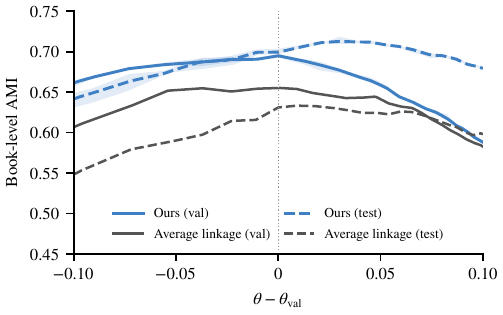}
\caption{Threshold sensitivity on Manga109. Solid and dashed lines show validation and test book-level AMI, respectively. For our method, lines show the mean over three models and shaded areas show their range. The horizontal axis is the offset from the threshold selected on validation.}
\label{fig:supp_threshold}
\end{figure}

\paragraph{Threshold sensitivity.}
\Cref{fig:supp_threshold} shows book-level AMI on the validation and test volumes as the inference threshold moves away from the value selected on validation.
Within $\pm0.05$ of the selected threshold, the test AMI of the three models ranges from 0.679 to 0.719, and every value exceeds the highest test AMI of average-linkage clustering over the whole range (0.633).
The test AMI of our method peaks at slightly higher thresholds than the selected ones, indicating that the reported results do not rely on a favorable threshold choice.

\begin{table}[t]
\centering
\small
\setlength{\tabcolsep}{3pt}
\caption{Selected epoch, learned coefficients, and validation-selected threshold $\theta$ of each model.}
\label{tab:supp_params}
\begin{tabular}{lccccccc}
\toprule
Seed & Epoch & $\alpha$ & $\gamma$ & $\epsilon$ & $a_p$ & $a_f$ & $\theta$ \\
\midrule
0 & 13 & $0.202$ & $-0.108$ & $1.059$ & $1.075$ & $1.059$ & $0.693$ \\
1 & 11 & $0.200$ & $-0.102$ & $1.059$ & $1.063$ & $1.060$ & $0.698$ \\
2 & 18 & $0.200$ & $-0.121$ & $1.059$ & $1.074$ & $1.058$ & $0.697$ \\
\bottomrule
\end{tabular}
\end{table}

\section{Additional Ablations}
\label{sec:supp_abl}

\Cref{tab:supp_abl} reports additional ablations on Manga109 that examine the positional kernel and the training objectives.

\paragraph{Panel distance.}
w/o panel distance removes the panel distance $u^{f}$ from the positional kernel, so that $\kappa$ depends only on page distance, while the same-panel mask and the same-panel term $Q$ are retained.
This change lowers book-level AMI from 0.699 to 0.684 and page-level AMI from 0.754 to 0.724.
Panel distance therefore provides positional evidence beyond whether two instances share a panel, although the page-level results of this variant vary considerably across seeds.

\paragraph{Training of the page-distance term.}
In our method, the page-distance term $g_{\psi}$ is learned only from the pair objective.
Joint training of $g$ additionally lets the merge objective update $g_{\psi}$, and w/o pair loss removes the pair objective so that all components, including $g_{\psi}$, are trained with the merge objective alone.
Both variants reduce performance: book-level AMI drops to 0.669 with joint training and to 0.610 without the pair loss, the latter falling below average-linkage clustering (0.631).
The decline grows as the merge objective takes a larger role in learning $g_{\psi}$.
This supports the design in \cref{sec:training}: merge candidates cover only a biased subset of page distances, so $g_{\psi}$ should be learned from all instance pairs.

\begin{table*}[t]
\centering
\small
\setlength{\tabcolsep}{5pt}
\setlength{\aboverulesep}{0pt}
\setlength{\belowrulesep}{0pt}
\renewcommand{\arraystretch}{1.18}
\caption{Additional ablations on Manga109. w/o panel distance uses only page distance in the positional kernel. Joint training of $g$ updates the page-distance term with both objectives. w/o pair loss trains all components with the merge objective; for this variant, the best trained epoch on validation is reported. Subscripts report sample standard deviations over three seeds.}
\label{tab:supp_abl}
\begin{tabular}{lcccccc}
\toprule
\multirow{2}{*}{Method}
& \multicolumn{4}{c}{Book-level}
& \multicolumn{2}{c}{Page-level} \\
\cmidrule(lr){2-5}
\cmidrule(lr){6-7}
& AMI $\uparrow$ & NMI $\uparrow$ & ARI $\uparrow$
& B$^3$ F1 $\uparrow$ & AMI $\uparrow$ & NMI $\uparrow$ \\
\midrule
\grouprow{bgabl}{7}{Additional ablations (Manga109)}
\rowcolor{labl}\ind Ours
& $0.699_{\pm0.004}$ & $0.727_{\pm0.003}$ & $0.601_{\pm0.006}$ & $0.677_{\pm0.001}$ & $0.754_{\pm0.002}$ & $0.836_{\pm0.001}$ \\
\rowcolor{labl}\ind w/o panel distance
& $0.684_{\pm0.012}$ & $0.709_{\pm0.012}$ & $0.581_{\pm0.024}$ & $0.663_{\pm0.013}$ & $0.724_{\pm0.035}$ & $0.809_{\pm0.040}$ \\
\rowcolor{labl}\ind Joint training of $g$
& $0.669_{\pm0.009}$ & $0.691_{\pm0.007}$ & $0.544_{\pm0.009}$ & $0.642_{\pm0.006}$ & $0.698_{\pm0.005}$ & $0.784_{\pm0.012}$ \\
\rowcolor{labl}\ind w/o pair loss
& $0.610_{\pm0.006}$ & $0.632_{\pm0.010}$ & $0.416_{\pm0.003}$ & $0.555_{\pm0.013}$ & $0.585_{\pm0.026}$ & $0.659_{\pm0.031}$ \\
\bottomrule
\end{tabular}
\end{table*}

\section{Per-Volume Results and Qualitative Examples}
\label{sec:supp_results}

\paragraph{Per-volume results.}
\Cref{tab:supp_pervol} reports book- or chapter-level AMI for each Manga109 test volume and each PopManga chapter.
On Manga109, our method outperforms average-linkage clustering on 10 of the 11 volumes and MagiV2 on all 11.
Averaged over volumes, the improvement is 0.068 over average-linkage clustering and 0.184 over MagiV2, with 95\% paired bootstrap intervals of $[0.038, 0.098]$ and $[0.144, 0.228]$, respectively; the three seeds are averaged within each volume before resampling volumes.
On PopManga, our method outperforms average-linkage clustering on 14 of the 20 PopManga-unseen chapters and 23 of the 30 PopManga-seen chapters, and outperforms MagiV2 on 20 and 29 of them, respectively.

\begin{table*}[p]
\centering
\footnotesize
\setlength{\tabcolsep}{3pt}
\setlength{\aboverulesep}{0pt}
\setlength{\belowrulesep}{0pt}
\renewcommand{\arraystretch}{1.1}
\caption{Book-level AMI on each Manga109 test volume and chapter-level AMI on each PopManga chapter. Models and thresholds are selected on Manga109. Subscripts report sample standard deviations over three seeds.}
\label{tab:supp_pervol}
\begin{minipage}[t]{0.49\linewidth}
\centering
\resizebox{\linewidth}{!}{%
\begin{tabular}[t]{lccc}
\toprule
Volume / chapter & Avg.\ link. & MagiV2 & Ours \\
\midrule
\grouprow{bgmanga}{4}{Manga109}
\rowcolor{lmanga}\ind ARMS & $0.664$ & $0.651$ & $0.798_{\pm0.009}$ \\
\rowcolor{lmanga}\ind AisazuNihaIrarenai & $0.599$ & $0.511$ & $0.581_{\pm0.006}$ \\
\rowcolor{lmanga}\ind AkkeraKanjinchou & $0.738$ & $0.480$ & $0.825_{\pm0.006}$ \\
\rowcolor{lmanga}\ind Akuhamu & $0.672$ & $0.552$ & $0.696_{\pm0.020}$ \\
\rowcolor{lmanga}\ind AosugiruHaru & $0.375$ & $0.263$ & $0.492_{\pm0.029}$ \\
\rowcolor{lmanga}\ind AppareKappore & $0.690$ & $0.455$ & $0.703_{\pm0.002}$ \\
\rowcolor{lmanga}\ind Arisa & $0.660$ & $0.531$ & $0.678_{\pm0.004}$ \\
\rowcolor{lmanga}\ind BEMADER\_P & $0.483$ & $0.403$ & $0.632_{\pm0.006}$ \\
\rowcolor{lmanga}\ind BakuretsuKungFuGirl & $0.601$ & $0.571$ & $0.693_{\pm0.009}$ \\
\rowcolor{lmanga}\ind Belmondo & $0.821$ & $0.722$ & $0.891_{\pm0.002}$ \\
\rowcolor{lmanga}\ind LoveHina\_vol14 & $0.637$ & $0.531$ & $0.705_{\pm0.008}$ \\
\grouprow{bgunseen}{4}{PopManga-unseen}
\rowcolor{lunseen}\ind Assassination Classroom, ch.~1 & $0.717$ & $0.632$ & $0.693_{\pm0.004}$ \\
\rowcolor{lunseen}\ind Assassination Classroom, ch.~2 & $0.808$ & $0.710$ & $0.750_{\pm0.009}$ \\
\rowcolor{lunseen}\ind Bleach, ch.~1 & $0.661$ & $0.570$ & $0.672_{\pm0.007}$ \\
\rowcolor{lunseen}\ind Bleach, ch.~2 & $0.739$ & $0.685$ & $0.775_{\pm0.012}$ \\
\rowcolor{lunseen}\ind Blue Exorcist, ch.~1 & $0.566$ & $0.525$ & $0.651_{\pm0.003}$ \\
\rowcolor{lunseen}\ind Blue Exorcist, ch.~2 & $0.528$ & $0.419$ & $0.569_{\pm0.053}$ \\
\rowcolor{lunseen}\ind Dr.~Stone, ch.~1 & $0.439$ & $0.371$ & $0.479_{\pm0.028}$ \\
\rowcolor{lunseen}\ind Dr.~Stone, ch.~2 & $0.820$ & $0.378$ & $0.858_{\pm0.000}$ \\
\rowcolor{lunseen}\ind Food Wars!, ch.~1 & $0.525$ & $0.550$ & $0.578_{\pm0.020}$ \\
\rowcolor{lunseen}\ind Food Wars!, ch.~2 & $0.825$ & $0.666$ & $0.782_{\pm0.010}$ \\
\rowcolor{lunseen}\ind Kagurabachi, ch.~1 & $0.509$ & $0.433$ & $0.574_{\pm0.003}$ \\
\rowcolor{lunseen}\ind Kagurabachi, ch.~2 & $0.553$ & $0.481$ & $0.593_{\pm0.006}$ \\
\rowcolor{lunseen}\ind Kuroko's Basketball, ch.~1 & $0.685$ & $0.635$ & $0.760_{\pm0.025}$ \\
\rowcolor{lunseen}\ind Kuroko's Basketball, ch.~2 & $0.775$ & $0.660$ & $0.782_{\pm0.032}$ \\
\rowcolor{lunseen}\ind Nisekoi, ch.~1 & $0.727$ & $0.656$ & $0.716_{\pm0.023}$ \\
\rowcolor{lunseen}\ind Nisekoi, ch.~2 & $0.725$ & $0.650$ & $0.773_{\pm0.014}$ \\
\rowcolor{lunseen}\ind Oshi no Ko, ch.~1 & $0.954$ & $0.854$ & $0.932_{\pm0.011}$ \\
\rowcolor{lunseen}\ind Oshi no Ko, ch.~2 & $0.702$ & $0.563$ & $0.675_{\pm0.050}$ \\
\rowcolor{lunseen}\ind World Trigger, ch.~1 & $0.747$ & $0.624$ & $0.780_{\pm0.002}$ \\
\rowcolor{lunseen}\ind World Trigger, ch.~2 & $0.639$ & $0.617$ & $0.739_{\pm0.013}$ \\
\bottomrule
\end{tabular}}
\end{minipage}\hfill
\begin{minipage}[t]{0.49\linewidth}
\centering
\resizebox{\linewidth}{!}{%
\begin{tabular}[t]{lccc}
\toprule
Volume / chapter & Avg.\ link. & MagiV2 & Ours \\
\midrule
\grouprow{bgseen}{4}{PopManga-seen}
\rowcolor{lseen}\ind Bakuman, ch.~1 & $0.709$ & $0.594$ & $0.742_{\pm0.010}$ \\
\rowcolor{lseen}\ind Bakuman, ch.~2 & $0.820$ & $0.563$ & $0.842_{\pm0.001}$ \\
\rowcolor{lseen}\ind Black Clover, ch.~1 & $0.621$ & $0.529$ & $0.706_{\pm0.018}$ \\
\rowcolor{lseen}\ind Black Clover, ch.~2 & $0.590$ & $0.535$ & $0.642_{\pm0.025}$ \\
\rowcolor{lseen}\ind Chainsaw Man, ch.~1 & $0.579$ & $0.478$ & $0.591_{\pm0.002}$ \\
\rowcolor{lseen}\ind Chainsaw Man, ch.~2 & $0.742$ & $0.689$ & $0.751_{\pm0.008}$ \\
\rowcolor{lseen}\ind Death Note, ch.~1 & $0.816$ & $0.712$ & $0.814_{\pm0.007}$ \\
\rowcolor{lseen}\ind Death Note, ch.~2 & $0.805$ & $0.876$ & $0.875_{\pm0.040}$ \\
\rowcolor{lseen}\ind Demon Slayer, ch.~1 & $0.494$ & $0.422$ & $0.529_{\pm0.060}$ \\
\rowcolor{lseen}\ind Demon Slayer, ch.~2 & $0.441$ & $0.314$ & $0.534_{\pm0.062}$ \\
\rowcolor{lseen}\ind Dragon Ball, ch.~1 & $0.678$ & $0.585$ & $0.696_{\pm0.009}$ \\
\rowcolor{lseen}\ind Dragon Ball, ch.~2 & $0.715$ & $0.184$ & $0.627_{\pm0.002}$ \\
\rowcolor{lseen}\ind Hell's Paradise, ch.~1 & $0.846$ & $0.709$ & $0.752_{\pm0.022}$ \\
\rowcolor{lseen}\ind Hell's Paradise, ch.~2 & $0.677$ & $0.678$ & $0.682_{\pm0.011}$ \\
\rowcolor{lseen}\ind Hunter x Hunter, ch.~1 & $0.737$ & $0.637$ & $0.765_{\pm0.013}$ \\
\rowcolor{lseen}\ind Hunter x Hunter, ch.~2 & $0.842$ & $0.807$ & $0.832_{\pm0.007}$ \\
\rowcolor{lseen}\ind JoJo's Bizarre Adventure, ch.~1 & $0.576$ & $0.510$ & $0.759_{\pm0.022}$ \\
\rowcolor{lseen}\ind JoJo's Bizarre Adventure, ch.~2 & $0.531$ & $0.384$ & $0.540_{\pm0.010}$ \\
\rowcolor{lseen}\ind Jujutsu Kaisen, ch.~1 & $0.638$ & $0.599$ & $0.648_{\pm0.003}$ \\
\rowcolor{lseen}\ind Jujutsu Kaisen, ch.~2 & $0.693$ & $0.609$ & $0.742_{\pm0.014}$ \\
\rowcolor{lseen}\ind My Hero Academia, ch.~1 & $0.599$ & $0.485$ & $0.633_{\pm0.017}$ \\
\rowcolor{lseen}\ind My Hero Academia, ch.~2 & $0.617$ & $0.370$ & $0.596_{\pm0.013}$ \\
\rowcolor{lseen}\ind Naruto, ch.~1 & $0.458$ & $0.323$ & $0.488_{\pm0.029}$ \\
\rowcolor{lseen}\ind Naruto, ch.~2 & $0.601$ & $0.388$ & $0.540_{\pm0.007}$ \\
\rowcolor{lseen}\ind One Piece, ch.~1 & $0.638$ & $0.551$ & $0.670_{\pm0.038}$ \\
\rowcolor{lseen}\ind One Piece, ch.~2 & $0.669$ & $0.594$ & $0.766_{\pm0.012}$ \\
\rowcolor{lseen}\ind Spy x Family, ch.~1 & $0.633$ & $0.493$ & $0.661_{\pm0.004}$ \\
\rowcolor{lseen}\ind Spy x Family, ch.~2 & $0.808$ & $0.626$ & $0.826_{\pm0.026}$ \\
\rowcolor{lseen}\ind Tokyo Ghoul, ch.~1 & $0.707$ & $0.579$ & $0.705_{\pm0.019}$ \\
\rowcolor{lseen}\ind Tokyo Ghoul, ch.~2 & $0.561$ & $0.412$ & $0.651_{\pm0.022}$ \\
\bottomrule
\end{tabular}}
\end{minipage}
\end{table*}

\begin{figure*}[p]
\centering
\includegraphics[width=\linewidth]{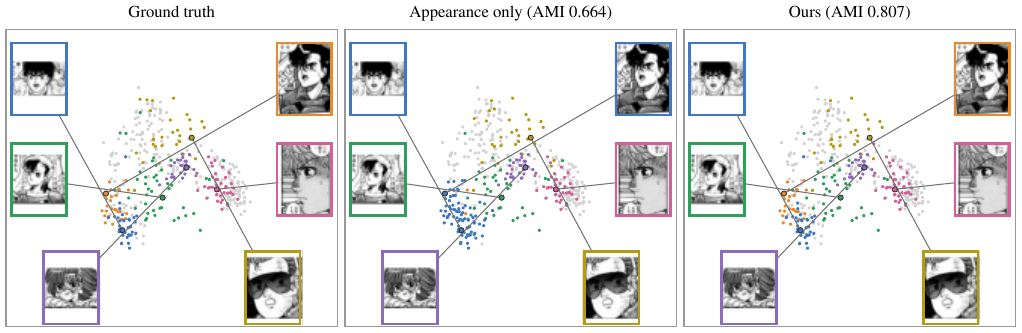}
\caption{Qualitative comparison on ARMS. From left to right: ground-truth identities, average-linkage clustering, and ours. All panels share the same two-dimensional coordinates. Six characters are highlighted, and each predicted cluster takes the color of its majority character; the remaining instances are shown in gray. AMI values are computed on the whole volume. Manga images are from ARMS \textcopyright Masaki Kato.}
\label{fig:supp_arms}

\vfill

\includegraphics[width=\linewidth]{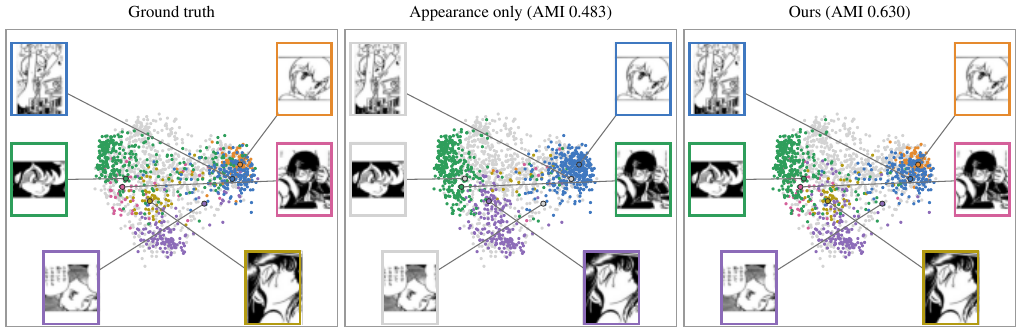}
\caption{Qualitative comparison on BEMADER\_P, visualized as in \cref{fig:supp_arms}. Manga images are from BEMADER\_P \textcopyright Yuichi Hasegawa.}
\label{fig:supp_bemader}

\vfill

\includegraphics[width=\linewidth]{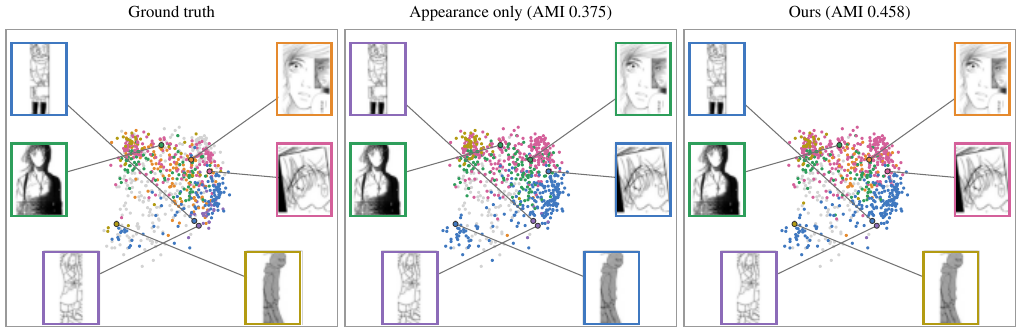}
\caption{Qualitative comparison on AosugiruHaru, visualized as in \cref{fig:supp_arms}. Manga images are from AosugiruHaru \textcopyright Momoko Okuda.}
\label{fig:supp_aosugiru}
\end{figure*}

\paragraph{Qualitative examples.}
\Cref{fig:supp_arms,fig:supp_bemader,fig:supp_aosugiru} show the three volumes with the largest improvements over average-linkage clustering.
They are visualized in the same way as \cref{fig:qualitative_comparison}, using the model trained with seed 0.
In each volume, average-linkage clustering merges several characters who look alike into the same cluster, whereas our method keeps them in separate clusters.

\begin{figure*}[p]
\centering
\includegraphics[width=\linewidth]{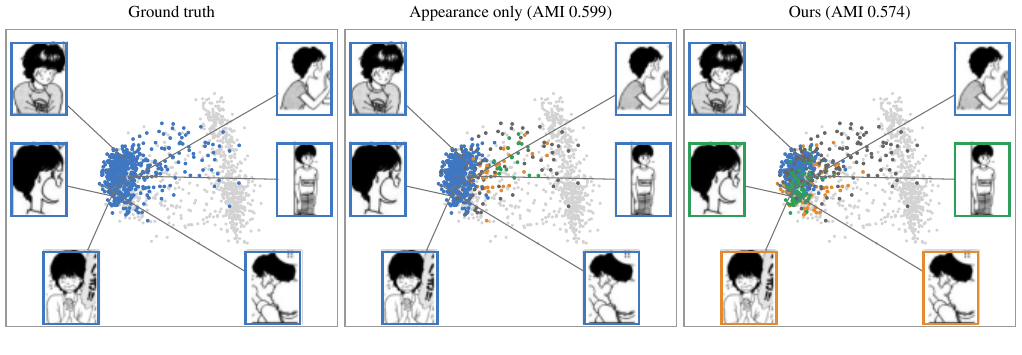}
\caption{Failure case on AisazuNihaIrarenai: fragmentation of one character. Only this character is highlighted, and the other characters are shown in light gray. In each prediction, blue, orange, and green mark the largest clusters containing this character, and dark gray marks its remaining clusters. The selected instances belong to one cluster under average-linkage clustering but are separated by our method. Manga images are from AisazuNihaIrarenai \textcopyright Masako Yoshi.}
\label{fig:supp_fail_aisazu}

\vfill

\includegraphics[width=\linewidth]{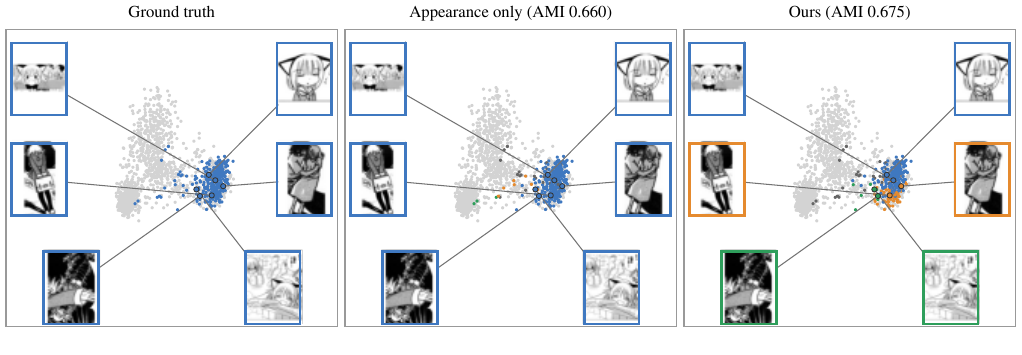}
\caption{Failure case on Arisa: fragmentation of one character in a volume whose overall score improves, visualized as in \cref{fig:supp_fail_aisazu}. Manga images are from Arisa \textcopyright Ken Yagami.}
\label{fig:supp_fail_arisa}

\vfill

\includegraphics[width=\linewidth]{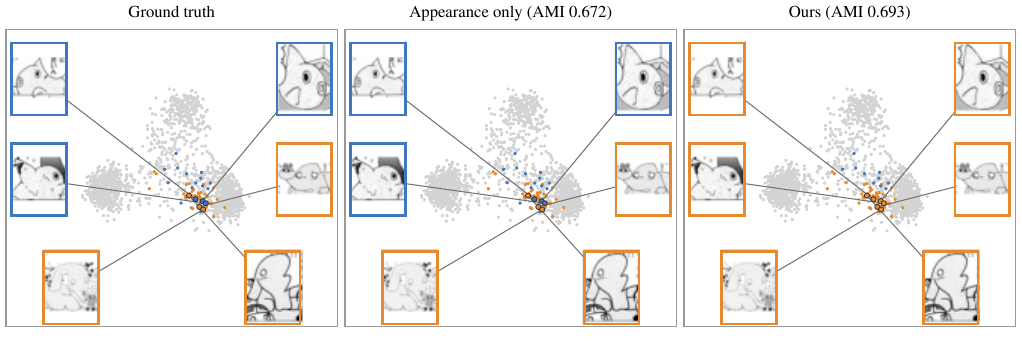}
\caption{Failure case on Akuhamu: merging of two different characters. Only these two characters are highlighted, and each predicted cluster takes the color of its majority character among them. Average-linkage clustering keeps the two characters apart, whereas our method merges them into one cluster. Manga images are from Akuhamu \textcopyright Satoshi Arai.}
\label{fig:supp_fail_akuhamu}
\end{figure*}

\Cref{fig:supp_fail_aisazu,fig:supp_fail_arisa,fig:supp_fail_akuhamu} show three types of failure.
On AisazuNihaIrarenai, the only test volume where our method scores lower than average-linkage clustering, a frequently appearing character is split into more clusters: instances that average-linkage clustering groups together are separated by our method.
On Arisa, a similar fragmentation of one character occurs even though the full-volume score improves, showing that gains at the volume level can coexist with local errors.
On Akuhamu, our method merges two visually similar characters that average-linkage clustering keeps apart.
Because merge decisions are irreversible, such errors persist once they are made, which motivates the revisable clustering discussed in \cref{sec:conclusion}.

\section{Merge Process}
\label{sec:supp_merge}

\begin{figure*}[p]
\centering
\includegraphics[width=\linewidth]{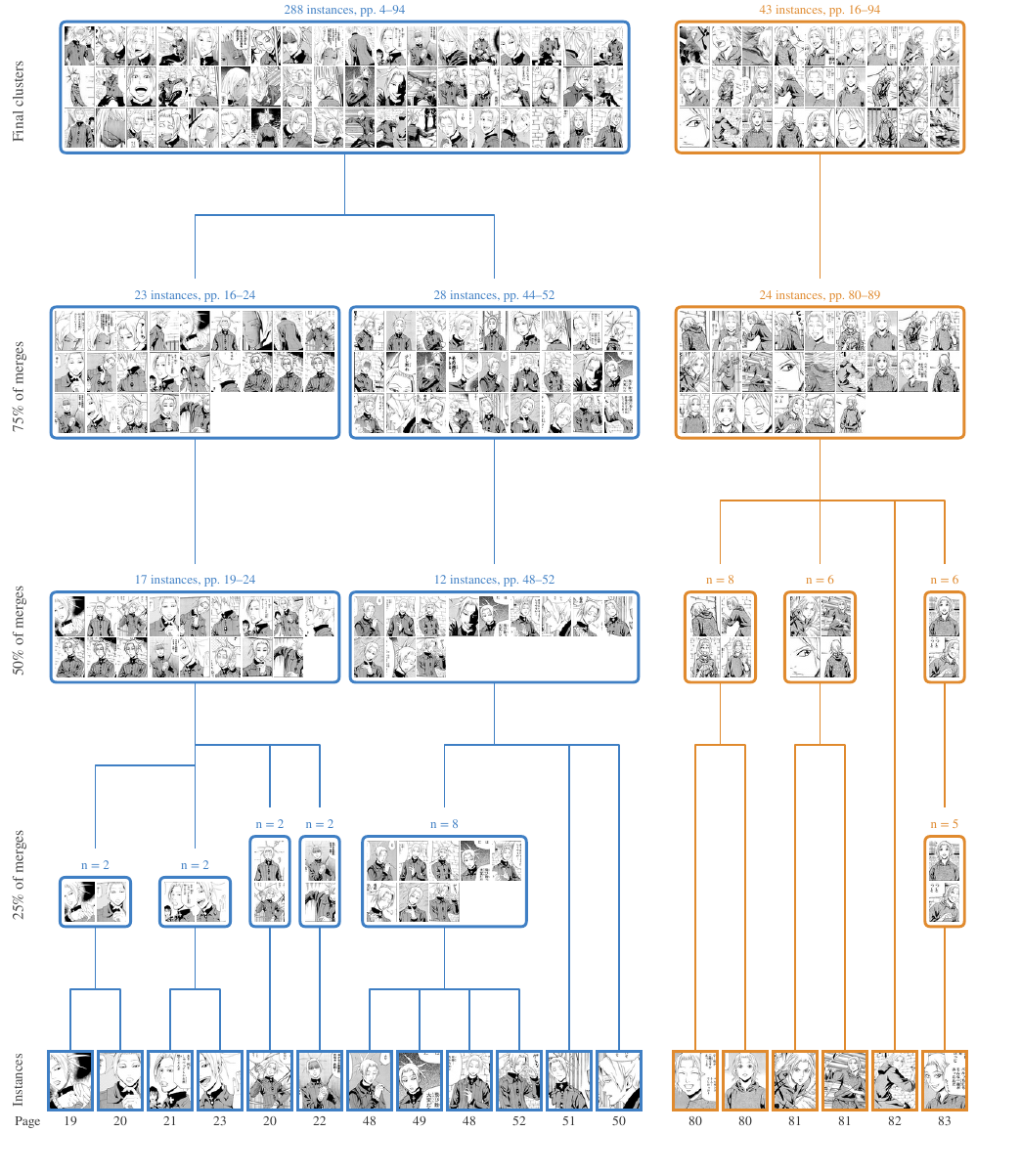}
\caption{Merge process of our method on the volume in \cref{fig:qualitative_comparison}. The bottom row shows selected instances of two characters (blue and orange) with their page numbers. Each box above shows the group containing these instances after 25\%, 50\%, and 75\% of the accepted merges and in the final clustering, together with its size and page range; lines indicate which groups are merged between stages. A line passing through a stage indicates an instance that has not yet been merged. Each box displays the members of the group, or a subset sampled uniformly over pages for large groups. Instances are ordered to follow the merge structure. Manga images are from Belmondo \textcopyright Shoei Ishioka.}
\label{fig:supp_merge}
\end{figure*}

\Cref{fig:supp_merge} illustrates how our method builds clusters on the volume shown in \cref{fig:qualitative_comparison}, using the model trained with seed 0.
We select instances of two visually similar characters from three page ranges and follow the groups that contain them through the actual merge sequence.
For each selected instance, the figure shows the group it belongs to after 25\%, 50\%, and 75\% of the accepted merges and in the final clustering.

The process proceeds from local to global.
After a quarter of the merges, the groups are still small and contain instances from the same or adjacent pages.
By the halfway point, these small groups have grown into larger groups, each covering a short range of pages.
Groups of the same character that lie far apart in the volume remain separate even after three quarters of the merges and are joined only in the final stage, forming a cluster that spans most of the volume.
Throughout the process, the two characters remain in separate clusters despite their similar appearance.

\end{document}